\documentclass[10pt,twocolumn,letterpaper]{article}

\usepackage{cvpr}
\usepackage{times}
\usepackage{epsfig}
\usepackage{graphicx}
\usepackage{amsmath}
\usepackage{amssymb}
\usepackage{booktabs}
\usepackage[breaklinks=true,bookmarks=false]{hyperref}

\cvprfinalcopy % *** Uncomment this line for the final submission

\def\cvprPaperID{****} % *** Enter the CVPR Paper ID here

\newcommand{\vvec}{\mathbf{v}}
\newcommand{\avec}{\mathbf{a}}

\begin{document}

%%%%%%%%% TITLE
\title{The YouTube-8M Kaggle Competition: Challenges and Methods}

\author{Haosheng Zou\thanks{Equal contribution.}\ \ \ \ \ Kun Xu\footnotemark[1]\ \ \ \ \ Jialian Li\ \ \ \ \ Jun Zhu\\
{\tt\small \{zouhs16, xu-k16, lijialia16\}@mails.tsinghua.edu.cn, dcszj@mail.tsinghua.edu.cn} \\
Department of Computer Science and Technology, Tsinghua University \\
Beijing, China
% \and
}

\maketitle
%\thispagestyle{empty}

%%%%%%%%% ABSTRACT
\begin{abstract}
We took part in the YouTube-8M Video Understanding Challenge hosted on Kaggle, and achieved the 10th place within less than one month's time. In this paper, we present an extensive analysis and solution to the underlying machine-learning problem based on frame-level data, where major challenges are identified and corresponding preliminary methods are proposed. It's noteworthy that, with merely the proposed strategies and uniformly-averaging multi-crop ensemble was it sufficient for us to reach our ranking\footnote{Our code is at \url{https://github.com/taufikxu/youtube} on branches kunxu and zhs.}. We also report the methods we believe to be promising but didn't have enough time to train to convergence. We hope this paper could serve, to some extent, as a review and guideline of the YouTube-8M multi-label video classification benchmark, inspiring future attempts and research.

\end{abstract}

%%%%%%%%% BODY TEXT
\section{Introduction}

Large-scale datasets such as ImageNet \cite{Deng2009ImageNet} play a key role in boosting modern machine learning \cite{Krizhevsky2012ImageNet}. Google has recently introduced the YouTube-8M dataset \cite{Abu2016YouTube}, the video equivalence to ImageNet, as an excellent testbed for general multi-label video classification. A competition based on the dataset was held on Kaggle \cite{kaggleweb} to encourage better models and results. Started code \cite{startercode} was also kindly provided.

% 数据集、评估也讲一讲

\subsection{Dataset Description}
\label{sec:data}
YouTube-8M consists of 7 million YouTube videos labeled with 4716 entities across diverse, general categories such as arts, books, games and food. Since directly processing the raw videos would be prohibitive at this scale, the data are already provided in the form of extracted visual and audio features from state-of-the-art perception models \cite{inceptionweb,45611}, at one-frame-per-second.

Both frame-level features and video-level features (uniformly averaged frame-level features within each video) are provided. The 4716 entities could be grouped into 25 high-level verticals including ``Unknown''. Labels for each video are generated by the YouTube video annotation system, and are not mutually exclusive across the 4716 entities, i.e., one video can have many corresponding labels (3.4 labels on average).

The videos are divided into 3 partitions for training, validation and testing. More detailed specifications of the dataset are given in Table \ref{tab:data}.

\begin{table}[htbp]
  \centering
  \caption{Dataset specs.}
    \begin{tabular}{|r|r|}
    \toprule
    Training set size & 4,906,660 \\
    Validation set size & 1,401,828 \\
    Test set size & 700,640 \\
    \midrule
    Visual feature dimension & 1,024 \\
    Audio feature dimension & 128 \\
    \midrule
    Feature sequence length & 120-360 \\
    \bottomrule
    \end{tabular}%
  \label{tab:data}%
\end{table}%

\subsection{Evaluation Metric}

The competition uses Global Average Precision (GAP) as the evaluation metric. Participants are required to output the top-20 predictions of each video in decreasing order of confidence, and the predictions of all the test videos are considered together. Denote the number of test videos as $N$, and the total number of positive labels in these $N$ videos as $M$, GAP is calculated as follows:

$$\mathrm{GAP} = \sum_{i=1}^{20N} \frac{p(i)}{i} \cdot \frac{1}{M},$$

where predictions of the whole test set are sorted in descending order of confidence, and $p(i)$ is the number of true positives in the first $i$ predictions.
To achieve a high GAP, all true positives should be given large confidence. Ideally, a submission with the first $M$ predictions after sorting corresponding to the $M$ positive labels achieves 1.0 GAP. As a result, although 20 predictions have to be made for each video, most of the 20 should be given low confidence so as not to rank ahead of true positives of other videos after sorting.

\section{Problem Definition}

A simple baseline result is already offered on Kaggle as ``mean\_rgb + mean\_audio Benchmark'' achieving 0.74711 public GAP. The provided mean-pooled video-level features are assumedly concatenated and fed directly to 4716 one-vs-all logistic regression classifiers, each predicting whether to annotate one label. Considering that video topics mostly appear at different positions within one video, such averaged features may not be a good descriptor of the whole video. Besides, directly working with the video-level features waves off most deep learning advances, as convolutional and recurrent neural networks are hardly applicable, leaving only MLP to be tried.
We therefore decided at the beginning of participation to focus on exploiting frame-level features, which is common practice for many other participating teams \cite{Miech2017Learnable,Wang2017The,fdt,skalic2017Deep}.

With most of our efforts devoted to frame modeling, we haven't exploited much the structure information between different labels, and simply view the multi-label classification problem as 4716 binary classification problems. More formally, our multi-label video classification problem is defined as: given the visual and audio feature sequences $\{\vvec_1, \vvec_2, ..., \vvec_T\}$ and $\{\avec_1, \avec_2, ..., \avec_T\}$ of a video as input, output the probability of positively labeling the video with entities $e_1, e_2, ..., e_{4716}$ respectively. $T$ is the sequence length, varying from video to video. As far as we know from the Kaggle discussion, many teams including some top ones consider the problem under this definition.

Towards this end, our model can be roughly divided into two blocks: \textbf{frame understanding block}, which extracts a fixed-length video descriptor $\mathbf{x}_{video}$ from frame-level features, and \textbf{classifiers block}, which performs classification based on the extracted video descriptor $\mathbf{x}_{video}$.

\section{Challenges}

The defined video classification problem has at least the following challenges:

\begin{enumerate}
\item \emph{Dataset Scale}: YouTube-8M is not only large-scale in terms of the number of data samples, but also in terms of the size of each data sample. Each video is between 120 and 500 seconds in length \cite{yt8mweb}, resulting in 120 to 360 1024-dimensional visual features (plus 128-dimensional audio) per video at a processing rate of 1 fps up to the first 360 seconds \cite{Abu2016YouTube} during feature extraction. Such individual size, combined with the total number of 5 million (or over 6 million if we include the validation set), poses remarkable challenge to training on a single GPU in terms of disk I/O and convergence rate. The GPU has to constantly stall to wait for data, and directly training the LSTM model provided by the starter code takes nearly a week to converge.

\item \emph{Noisy Labels}: % 之后的解决： 用crop？和ensemble
The labels are machine-generated by the YouTube video annotation system, rather than crowd-sourcing from human workers as ImageNet \cite{Deng2009ImageNet}. The annotation system is in essence a rule-based system, based on metadata, context and query click signals, etc. \cite{Abu2016YouTube}, hence far-from-human-level labeling. The labels are quantified on a limited subset of 8000 videos to have 78.8\% precision and 14.5\% recall w.r.t. manual labeling, suggesting that plenty of positive entities are instead labeled negative. As a result, negative labels dominates most entities; it's not very reliable to analyze the prediction results on the training set; and we are more like machine-learning the annotation system unnaturally than learning a perception system of humans.

\item \emph{Lack of Supervision}: % 之后的解决： attention, 没跑完的AE
The ground-truth values are only video-level, and we don't have any information for each frame, whether the location of each label within the video or the relative prominence of each frame. Therefore, the video-level supervision signals provided by the video-level ground truths are supposed to guide the learning of not only the classifiers block but also the frame understanding block (with frame number at the scale of hundreds), casting doubts on the effectiveness of the supervision signals on learning the whole model.

\item \emph{Temporal Dependencies}: % 之后的解决： 视频时序
The visual features are extracted by feeding one frame's image per second to a powerful pre-trained image recognition model \cite{inceptionweb}, so the features haven't yet taken into account the temporal dependencies across frames lying in the nature of videos. Though the images are taken one-frame-per-second, we humans still can usually identify most video topics watching a video at such frame rate, but may not be able to do so if the frame order are randomized. Our frame understanding block is therefore supposed to model such temporal dependencies.
Top teams have resorted to RNNs \cite{Miech2017Learnable,Wang2017The,fdt,skalic2017Deep} (such as LSTM \cite{Hochreiter1997LongSM} and GRU \cite{Cho2014On}) and clustering-based methods \cite{Miech2017Learnable,fdt} (such as generalized VLAD \cite{Arandjelovi2015NetVLAD} and bag-of-visual-words \cite{Philbin2008Lost}) in this regard.

\item \emph{Multi-modal Learning}:
\label{modal}
Audio features are provided along with the visual features in the updated version of the dataset \cite{yt8mweb}. Although the dataset was originally constructed following the principle that ``every label in the dataset should be distinguishable using visual information alone'' \cite{Abu2016YouTube}, it's fair to expect that audio features will help classification, which is shown by experiments detailed later in Table \ref{tab:res}.
We humans generally watch most videos with audio turned on, and, for example, entity ``concert'' and ``motorsport'' surely sound distinctly. Including audio features constitutes a multi-modal learning problem \cite{Ngiam2012Multimodal}, and the challenge then arises as how to combine and exploit the two-modal features.

\item \emph{Multiple Labels}: % 之后的解决： 25个网络？
One video can have multiple labels, and, in fact, have 3.4 labels on average, which is different from ImageNet classification with mutually exclusive labels where each input belongs to exactly one output class. This is also reflected in that we view the problem as 4716 binary classification tasks rather than one 4716-class softmax classification task. The most straightforward pipeline of building one frame understanding block followed by 4716 binary classifiers is thus debatable because the uniquely extracted feature $\mathbf{x}_{video}$ is to be used for 4716 different binary classification problems, which requires the feature to be incredibly descriptive.

Moreover, it may not be appropriate to treat the 4716 binary classification problems as independent with each other, as topics within one regular video are usually present or not present in groups, rather than independently, forming certain structures between the labels. 
Sharing the video understanding block, the classifiers are \emph{implicitly} correlated, but this may not be enough to model the correlations between labels. The top-2 teams have proposed Context Gating \cite{Miech2017Learnable} and Classifiers Chaining \cite{Wang2017The,Read2011Classifier} to deal with the challenge. Further inspirations could be drawn from devoted multi-label classification literature.

\item \emph{In-class Imbalance}:
Partly due to the low recall of the machine-generated labels, the ratio between positive and negative labels within almost all of the 4716 classes is highly imbalanced. More specifically, only 3 entities have more than 500K positive labels in the training set according to the vocabulary csv file \cite{yt8mweb}; less than 400 entities have more than 100K; and about 1000 entities have only hundreds. Therefore, 90\% of the binary classification problems have an imbalance ratio of under $\frac{100K}{5M} = \frac{1}{50}$, which are already extremely imbalanced, making it rather difficult to adequately learn each entity.
\end{enumerate}

\section{High-Level Algorithms}

\subsection{Prototype}
\label{sec:proto}
An LSTM + MoE (Mixture of Expert \cite{Abu2016YouTube}) model is provided in the starter code. Visual and audio features of each frame are first concatenated as input.
The model sequentially feeds all the frame features to a stacked LSTM as the frame understanding block, and uses the last output of the top-layer LSTM as the video descriptor for the 4716 classifiers, as depicted in Figure \ref{fig:model} without all the further modifications stated in the parentheses. We refer to this model as \emph{prototype}.

Due to the video length (around 225 frames on average \cite{Abu2016YouTube}), training for 5 epoches on the training set alone takes nearly one week at a traversing speed of 40 examples per second on our GeForce Titan X GPU, possibly limited by the I/O of our hard disk. 
During training we would only observe GAP on the current minibatch, and unfortunately GAP evaluation of this model on the validation set takes about prohibitively 10 hours, hence almost impossible to tune hyperparameters and compare model designs based on validation performance given the one month's time we had. These are all challenges posed by the \emph{dataset scale}.

We therefore make a compromise to train directly on both training and validation sets, and treat the public score as validation performance. Bearing this in mind, we try not to submit often, which is reflected in our few total submissions. We've only submitted 31 entries to Kaggle, which is the third least among the top-10 teams.

A better solution, however, would be to leave out a much smaller validation set, (for example, 20K videos, as done by the top-2 teams \cite{Miech2017Learnable,Wang2017The}), so that relatively efficient and reliable validation and model selection for ensemble could be done locally.

\subsection{Random Cropping}
\label{sec:crop}
Motivated mainly by the long training time as well as the concern that lengths at the scale of hundreds may be too difficult for an LSTM to capture the \emph{temporal dependencies}, we apply random cropping to the sequence input inspired by the random-cropping-patches widely adopted in image classification \cite{Krizhevsky2012ImageNet}. We first make further compromise to consider only up to the first 225 frames, and then down sample 5 times to 45 frames. More specifically, during the training phase, only the first frame is picked randomly from the first 10 frames, and later frames are picked deterministically by incrementing the frame index by stride 5.

The rationale behind this is that, humans could still get a grasp of the video topics looking at a few sampled frames from the video in temporal order. 
This is also supported by the only 0.0002 GAP decrease when training a same model with random cropping, as will be detailed in Table \ref{tab:res}.
Note that we still preserve the \emph{temporal dependencies} across frames unlike the DBoF model proposed in \cite{Abu2016YouTube}. Decreasing the sequence length by an order of magnitude, we are able to reach a traversing speed of 100 examples per second. The number of data samples is equivalently increased by an order of magnitude, but we didn't observe any performance gains by training the model for more than 5 epoches. % TODO: Therefore, such random crop is not to be treated as data augmentation.

By reducing the size of each data sample with simple random cropping, we mitigate the challenge of \emph{dataset scale}. It would also be expected that because of such \emph{noisy labels} in this dataset, the additional stochasticity introduced by random cropping would not harm performance much, which is already shown empirically. The relative \emph{lack of supervision} is in a way alleviated by reducing the depth (in time) of the frame understanding block.

\subsection{Multi-Crop Ensemble}
\label{sec:ensemble}
The aforementioned random cropping operation has a pleasant by-product: we could get multiple different predictions with one very model by varying the start index of each video crop. Cheap ensemble could then be performed without training different models, but instead with different views of the same video as input, as the common multi-crop evaluation in image classification \cite{Szegedy2016Rethinking}. Ensemble is shown to be one of the solutions to \emph{noisy labels} \cite{Melville2004Experiments}, increasing robustness of the classifiers.

During the test phase, we employ the simplest form of ensemble, something like bagging, where different predictions are averaged. We make five different top-20 predictions from each model by manually specifying the start index to be from 0 to 4 (or randomly sampled between 0 and 4, to get a single prediction without ensemble). Labels not in the top-20 are treated with confidence zero, the different predictions of different start indices and different models are then uniformly averaged altogether, and the averaged top-20 predictions are submitted at last. Under the GAP evaluation, the predicted confidence score matters a lot where positive labels should be given as high confidence as possible and negative as low as possible. By averaging the predictions, confidence scores are strengthened when different predictions share similar ideas, and weakened when they differ. The label noise would also be eliminated to some extent. Empirical results show notable performance gain.

\subsection{Early Stopping}

% train first, observe overfitting on validate and test
% 5 epoches fixed

The prototype model gives us 0.80224 public GAP after 5 epoches of training, while outputting GAPs always above 0.83 on the last thousands of minibatches. To investigate the probable overfitting, we evaluate the training set GAP to be 0.836 and the validation set GAP to be 0.802, suggesting certain overfitting (which is notable enough, considering the dataset scale and the few training epoches). We trained the prototype for more epoches and, as expected, observed increasingly degraded public GAP. However, earlier checkpoints before 5 epoches such as 3 and 4.5 epoches didn't give better performance either, though we didn't experiment intensively with earlier checkpoints.

Based on the above observation, 5 epoches seems an adequate training amount. In view of the long validation time (Section \ref{sec:proto}),
we therefore adopt early stopping to cope with the absence of validation and to try to prevent overfitting, i.e., we only train for 5 epoches on both training and validation sets. We fix this stopping criterion for all later models.

\section{Models}
\label{sec:models}
\subsection{Stacked LSTM + LR / MoE}
\label{sec:general}
% 1. data augmentation,,, 2. ensemble,,, maybe some analysis TODO 3. layer norm,,, 4. attention,,, 5. late fusion,,,

\begin{figure}
\centering
\includegraphics[width=0.99\columnwidth]{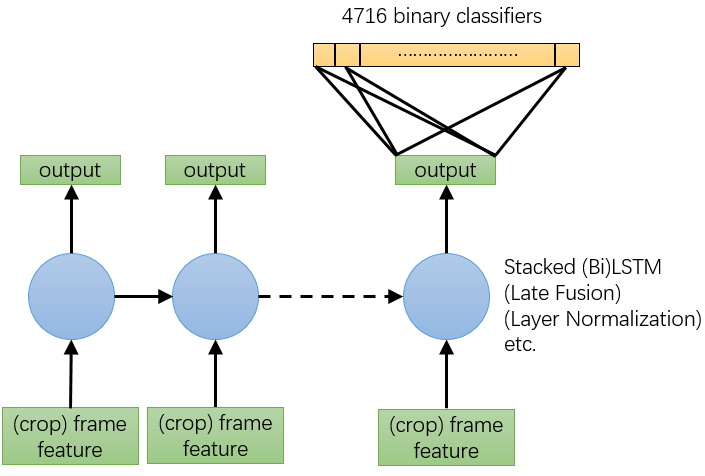}
\caption{General illustration of the models.}
\label{fig:model}
\end{figure}

Our general pipeline follows the prototype LSTM model provided by the starter code \cite{startercode}, as shown in Figure \ref{fig:model}. It only differs from the prototype in that it takes cropped sequence as input.

More specifically, an LSTM is used to process all the cropped, concatenated frame features as the frame understanding block, and the final output is used as the video descriptor for the classifiers, which we simply use 4716 logistic regressions (LRs) or the Mixture of Expert (MoE, 2 LR experts) model suggested in \cite{Abu2016YouTube}. We stack two layers of 1024 LSTM cells, and the final output of the top layer is extracted.
The gradients of the classification loss is back-propagated to the LSTM weights. In other words, the frame understanding LSTM is trained \emph{discriminatively} from the last state backwards to extract video descriptors.

Recent deep learning advances demonstrate the scalability of RNNs in sequence tasks \cite{Graves2013HybridSR,Xu2015ShowAA,Bahdanau2014NeuralMT} as well as the success of LSTMs \cite{Hochreiter1997LongSM} in capturing long-term \emph{temporal dependencies}, which persuades us to apply LSTMs in all our models' frame understanding block.
Although more sophisticated classifiers could be adopted such as MLPs, we stick to simple LRs or their mixture. The considerations are that the frame understanding LSTM is much more difficult to train than the classifiers, so we shouldn't add more burden to the whole model by incorporating MLPs, and also that the discriminative supervision signals for the LSTM coming from the classifiers would make better sense if the classifiers are shallow, well-behaved and even linear models.

% The LSTM frame understanding block shows significant improvement over DBoF models, capturing more \emph{temporal dependencies} across frames.

\subsection{Layer Normalization}

Any LSTM-based model with random cropping (for example, Section \ref{sec:general}), more than doubling the traversing speed, took 3 to 4 days to train for 5 epoches regardless of the further modifications we made to the model. We were still not satisfied with this convergence rate given the relatively little time we had. 
It was only a few days ahead of the competition deadline that we realized to try Layer Normalization \cite{Ba2016LayerN}, one of the recurrent versions of Batch Normalization \cite{ioffe2015batch}, in our LSTM cells. Prior to that, we had only used the most basic LSTM cell without cell clipping, projection layers or peep-hole connections in our models.

Surprisingly, with Layer Normalization, the training GAP on minibatches quickly reaches 0.80 in less than 1 epoch, while almost maintaining the traversing speed. After about merely 2.7 epoches of training (at the last day when we have to stop to run inference), a single two-layer bidirectional (detailed later) LSTM model achieves 0.80736 public GAP, approaching the performance of a similar model without Layer Normalization trained for 5 whole epoches. We have to leave it for future work to fully explore the capability of layer-normalized LSTM cells, and our preliminary results demonstrate the high potential of the technique in further reducing training time and overcoming the challenging \emph{dataset scale}.

\subsection{Attention}

Set out to tackle the \emph{lack of supervision}, we wish to apply supervision loss not restricted to the last state of the LSTM. Although LSTMs have the capability of keeping a ``long-term memory'', it's doubtful that the model in Figure \ref{fig:model} will be anywhere near sensitive to early frames of videos. We humans, however, can usually tell several topics of a video from the first few seconds.
It would be desired that the model could more explicitly attend to and extract feature from all the frames.

\begin{figure}
\centering
\includegraphics[width=0.99\columnwidth]{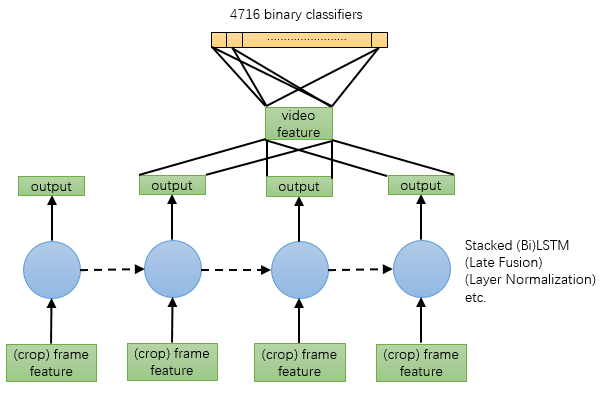}
\caption{Our model with attention.}
\label{fig:model_att}
\end{figure}

We therefore draw inspiration from the attention mechanism in neural machine translation \cite{Bahdanau2014NeuralMT}. Instead of learning an adaptive attention, we implement a much simplified, pre-specified version, where the one-third, two-thirds and the last outputs of the LSTM are given equal attention and mean-pooled into the video descriptor, as depicted in Figure \ref{fig:model_att}. With this model, in the meantime of capturing the \emph{temporal dependencies} (the three segments are not treated as separate video inputs), supervision is able to be injected earlier into the LSTM to ease training, leaning the sensitivity of the LSTM from the last few frames to all.

We've considered adaptive attention, where the three attention weights are instead determined by a neural network. We didn't favor complex networks, but also held that one fully-connected layer (equivalent to linear regression) upon the three features doesn't make much sense, since it's unrealistic to determine each importance with only one feature direction (the regression weight). So we've only experimented with the simplest form of attention.

\subsection{Bidirectional LSTM}

\begin{figure}
\centering
\includegraphics[width=0.8\columnwidth]{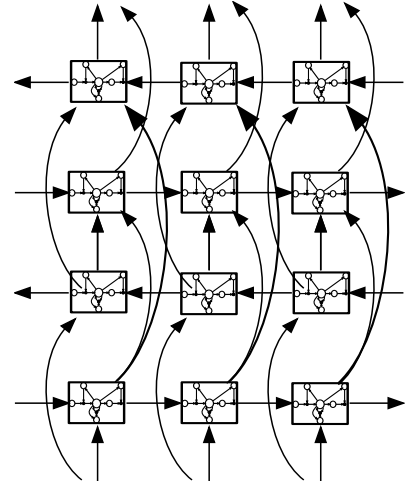}
\caption{Stacked BiLSTM model \cite{Graves2013HybridSR}.}
\label{fig:bilstm}
\end{figure}

Bidirectional LSTMs (BiLSTMs), as shown in Figure \ref{fig:bilstm}, have proved successful in speech recognition \cite{Graves2013HybridSR} and have become the standard model in that area \cite{Zeyer2016ACS}. We exploit the stronger capability of stacked BiLSTMs in capturing \emph{temporal dependencies} across frames and context modeling at each frame. It's not most natural for video understanding since we humans generally do not watch videos backwards, but we argue that by incorporating the information from both the past and future frames at each timestep, the RNN could learn better representations for each frame, hence for the whole video. The rationale is two-fold: on one hand, the top-layer RNN would be presented with the same amount of information (all the frames) at each timestep, while with vanilla RNN the information increases incrementally in an auto-regressive manner; on the other hand, the final output of the top layer would be the concatenation of the final outputs of the two directions, improving the sensitivity of the video descriptor to earlier frames.

As for implementation, we stack two bidirectional LSTMs on top of the frame features to extract video descriptors, with the first layer of 1024 units in each direction and the second layer of \{768, 1024\} units.

\subsection{Late Fusion}

The most straightforward way to exploit the provided visual and audio features is to concatenate them together to form a 1152-dimensional feature as the input to LSTM, as is done in all previous models. However, the provided 1024-dimensional video features and 128-dimensional audio features are extracted from different networks (\cite{inceptionweb} for visual and \cite{45611} for audio features), and it should be noted that the visual features haven't taken into account the \emph{temporal dependencies} yet. The two modals of features may not, therefore, represent the same level of information semantically. 

In this regard, we propose to add one layer of LSTM on either of the features separately and then concatenate the LSTM outputs as inputs to the next layer, as a practice of late fusion in \emph{multi-modal learning}. The model is able to learn different dynamics within video and audio independently (as we humans have separate low-level perception systems for visual and audio inputs), and then concatenate the features at a more similar semantic level.

\section{Results}

% We have this prototype model trained for 5 epoches and...

We took part in the competition less than a month ago ahead of the deadline, with our first submission made on May 5th. Consequently, the methods stated in Section \ref{sec:models} were developed incrementally, and we were not able to train a single final model incorporating them all. We present our results in Table \ref{tab:res} of the models we've submitted, each with different modifications on the prototype model. We didn't tune the hyperparameters, except specifying the base learning rate to be 0.001,number of epoches to be 5 and batchsize to be 128.

\begin{table}[htbp]
  \centering
  \caption{Leaderboard results in GAP. The first four models are trained on the training set only. Prototype is the LSTM+MoE model stated in Section \ref{sec:proto}. Models without ``full'' are trained with random cropping. LN stands for Layer Normalization. Visual and audio features are directly concatenated at input except the late-fusion model. All results except ensembles are from a single prediction of a single model, with start index randomly sampled (Section \ref{sec:ensemble}).}
    \begin{tabular}{|l|r|r|}
    \toprule
    Model & \multicolumn{1}{l|}{Public} & \multicolumn{1}{l|}{Private} \\
    \midrule
    baseline (on Kaggle) & 0.74711 & 0.74714 \\
    prototype (full, visual only) & 0.78105 & 0.78143 \\
    prototype (full) & 0.80224 & 0.80207 \\
    prototype (crop) & 0.80204 & 0.80190 \\
    \midrule
    BiLSTM+LR+LN & 0.80761 & 0.80736 \\
    BiLSTM+MoE & 0.81055 & 0.81067 \\
    BiLSTM+MoE+attention & 0.81232 & 0.81227 \\
    BiLSTM+MoE (full) & 0.81401 & 0.81399 \\
    ENSEMBLE (16) & 0.83477 & 0.83470 \\
    \textbf{ENSEMBLE (36)} & \textbf{0.83670} & \textbf{0.83662} \\
    \bottomrule
    \end{tabular}%
  \label{tab:res}%
\end{table}%

As can be seen from Table \ref{tab:res}, frame-level prototype models easily surpass the baseline built upon mean-pooled visual and audio features as expected, though it's reported by some teams \cite{skalic2017Deep} that video-level MLPs based on the mean-pooled features could achieve similar gain through certain feature engineering. The prototype model without audio features (visual only) degrades in performance, verifying the use of audio features and \emph{multi-modal learning}. The difference between cropped prototype and full prototype is negligible, validating our random cropping suggested in Section \ref{sec:crop}. The cropped prototype serves as baseline for later models.

Improvement brought by BiLSTM, Layer Normalization and attention could be observed from the second half of Table \ref{tab:res}, though it may not be considered so significant numerically, also because BiLSTM nearly doubles the number of parameters w.r.t. vanilla LSTM, and the second-half models are trained together with validation set.

Fundamental improvement is achieved from ensemble of the different predictions and models we've trained, which is a somewhat surprise even to ourselves since we didn't conduct any ensemble until the last day of the competition. Averaging 16 different predictions boosts GAP by 0.02, where the different predictions come from only four models (BiLSTM+LR+LN, BiLSTM+MoE, BiLSTM+MoE+attention, full BiLSTM+MoE) varying the start index as detailed in Section \ref{sec:ensemble}.
Our final result is the ensemble of 36 different predictions from more models (additionally, LSTM+late fusion+LR, LSTM+LR+LN, which are not submitted individually) and more start indices (5, 6, 7). The final prediction reaches 0.83662 private GAP. The success of ensemble suggests the diversity of the different models we proposed and the power of each individual model to some extent. Note that we only used the last checkpoint of each model for inference, rather than different checkpoints of the same model \cite{fdt,skalic2017Deep}.

It's worth mentioning that even though cropping accelerated training to 100 examples per second, we still observed that the GPU volatile utility (from the ``nvidia-smi'' command) drops to 0\% for thirty to forty percent of one minibatch's time.  We suspect it's because the GPU has to stall to wait for data input. With sufficient number of readers (we never used less than 6), we suppose it is due to the I/O speed of our hard disk, implying that hardware is also one of the solutions to \emph{dataset scale}.

\section{Other Methods}

% 4. loss manipulation NOT working 5. representation learning not tested... 6. 25 lstms too slow 
% 这部分毕竟应该略写的

We briefly introduce other methods we believe to be promising, but didn't have enough time and resources to explore or train to convergence.

\subsection{Separating Tasks}

For the challenge of \emph{multiple labels}, we argue that one unique descriptor for one video (of dimension from 1024 to 2048 in our models) may not be rich enough for 4716 classification tasks, since 2048 is still less than half of 4716. We therefore propose to divide the tasks into 25 meta-tasks according to the 25 high-level verticals as in Section \ref{sec:data}, and have different frame understanding LSTMs for each meta-task. Each LSTM extracts video descriptors for each meta-task discriminatively and differently. However, this model is huge and too slow to train (traversing speed at only 15 examples per second). It demonstrates partial potential reaching 0.60+ GAP in only 10K minibatches during training. We should have considered grouping the tasks further into 5 or 6 higher-level tasks though.

\subsection{Loss Manipulation}

For the challenge of \emph{in-class imbalance}, we tried ignoring negative labels when their predicted confidences are less than 0.15, but found it leading to worse GAP. Perhaps 0.15 is too large a threshold, since negative labels should also be effectively suppressed in order to reach a high GAP. We may also try random undersampling of the dominant negative data samples, or other techniques dealing with imbalanced data \cite{he2009learning}.

\subsection{Unsupervised Representation Learning}

For the challenge of \emph{lack of supervision}, we propose to conduct unsupervised learning first, i.e., using visual features to reconstruct both visual and audio features (the video-only auto-encoder architecture in \cite{Ngiam2012Multimodal}), so that we would have rich supervision every timestep from regression loss. The middle hidden layer activations could be used as video descriptor. We didn't have enough time to train this model.

\section{Conclusion}

We extensively analyze the challenges underlying the YouTube-8M multi-label video classification problem based on frame features, namely dataset scale, noisy labels, lack of supervision, temporal dependencies, multi-modal learning, multiple labels and in-class imbalance. Preliminary methods are proposed to mitigate the difficulties such as ensemble, attention and BiLSTM. These simple techniques have proved effective by our 10th final ranking on Kaggle. We expect even better results, given more time to fully exploit the methods and finish our other ideas.

\section*{Acknowledgements}
This work was supported by the National Basic Research Program of China (2013CB329403), the National Natural Science Foundation of China (61620106010, 61621136008) and the grants from NVIDIA and the NVIDIA DGX-1 AI Supercomputer.

%==============================================================================
%==End of content==============================================================
%==============================================================================

{\small
\bibliographystyle{ieee}
\bibliography{egbib}
}

\end{document}